\documentclass[runningheads]{llncs}
\usepackage{xcolor}
\usepackage{graphicx}
\usepackage{amsmath}
\usepackage{amssymb}
\usepackage{stmaryrd}
\usepackage{multirow}
\usepackage{booktabs}
\usepackage{todonotes}
\usepackage[hidelinks]{hyperref}

\newcommand\blfootnote[1]{%
  \begingroup
  \renewcommand\thefootnote{}\footnote{#1}%
  \addtocounter{footnote}{-1}%
  \endgroup
}

\begin{document}
\title{Spot What Matters: Learning Context\texorpdfstring{\\}{}
Using Graph Convolutional Networks\texorpdfstring{\\}{} 
for Weakly-Supervised Action Detection}
\titlerunning{Learning Context Using GCN for Weakly-Supervised Action Detection}
%


\author{Michail Tsiaousis\inst{1\thanks{Corresponding author}, \thanks{This work was carried out during an internship at TNO}}\orcidID{0000-0002-9355-5026} \and Gertjan Burghouts\inst{2}\orcidID{0000-0001-6265-7276} \and
Fieke Hillerstr\"{o}m\inst{2}\orcidID{0000-0003-1301-3073} \and
Peter van der Putten\inst{1}\orcidID{0000-0002-6507-6896}}

\authorrunning{M. Tsiaousis et al.}

\institute{Leiden University, Niels Bohrweg 1, 2333 CA Leiden, The Netherlands \and
TNO, Oude Waalsdorperweg 63, 2597 AK The Hague, The Netherlands\\
\email{tsiaousis.michail@gmail.com, \{gertjan.burghouts,fieke.hillerstrom\}@tno.nl, p.w.h.van.der.putten@liacs.leidenuniv.nl}}

\maketitle              
\begin{abstract}
The dominant paradigm in spatiotemporal action detection is to classify actions using spatiotemporal features learned by 2D or 3D Convolutional Networks. We argue that several actions are characterized by their context, such as relevant objects and actors present in the video. To this end, we introduce an architecture based on self-attention and Graph Convolutional Networks in order to model contextual cues, such as actor-actor and actor-object interactions, to improve human action detection in video. We are interested in achieving this in a weakly-supervised setting, i.e. using as less annotations as possible in terms of action bounding boxes. Our model aids explainability by visualizing the learned context as an attention map, even for actions and objects unseen during training. We evaluate how well our model highlights the relevant context by introducing a quantitative metric based on recall of objects retrieved by attention maps. Our model relies on a 3D convolutional RGB stream, and does not require expensive optical flow computation. We evaluate our models on the DALY dataset, which consists of human-object interaction actions. Experimental results show that our contextualized approach outperforms a baseline action detection approach by more than 2 points in Video-mAP. Code is available at \url{https://github.com/micts/acgcn}.

\keywords{Weakly-Supervised action detection  \and graph convolutional networks \and relational reasoning \and actor-context relations}
\end{abstract}
\section{Introduction}

\blfootnote{Paper presented at the International Workshop on Deep Learning for Human-Centric Activity Understanding (DL-HAU2020), January 11, 2021}

Human action recognition is an important part of video understanding, with potential applications in robotics, autonomous driving, surveillance, video retrieval and healthcare. Given a video, spatiotemporal action detection aims to localize all human actions in space and time, and classify the actions being performed. The dominant paradigm in action detection is to extend CNN-based object detectors \cite{10.1007/978-3-319-46448-0_2,7485869} to learn appearance and motion representations in order to jointly localize and classify actions in video. The desired output are action tubes \cite{Gkioxari2015FindingAT}: sequences of action bounding boxes connected in time throughout the video.

\begin{figure}[tb]
    \centering
    \includegraphics[width=.8\textwidth]{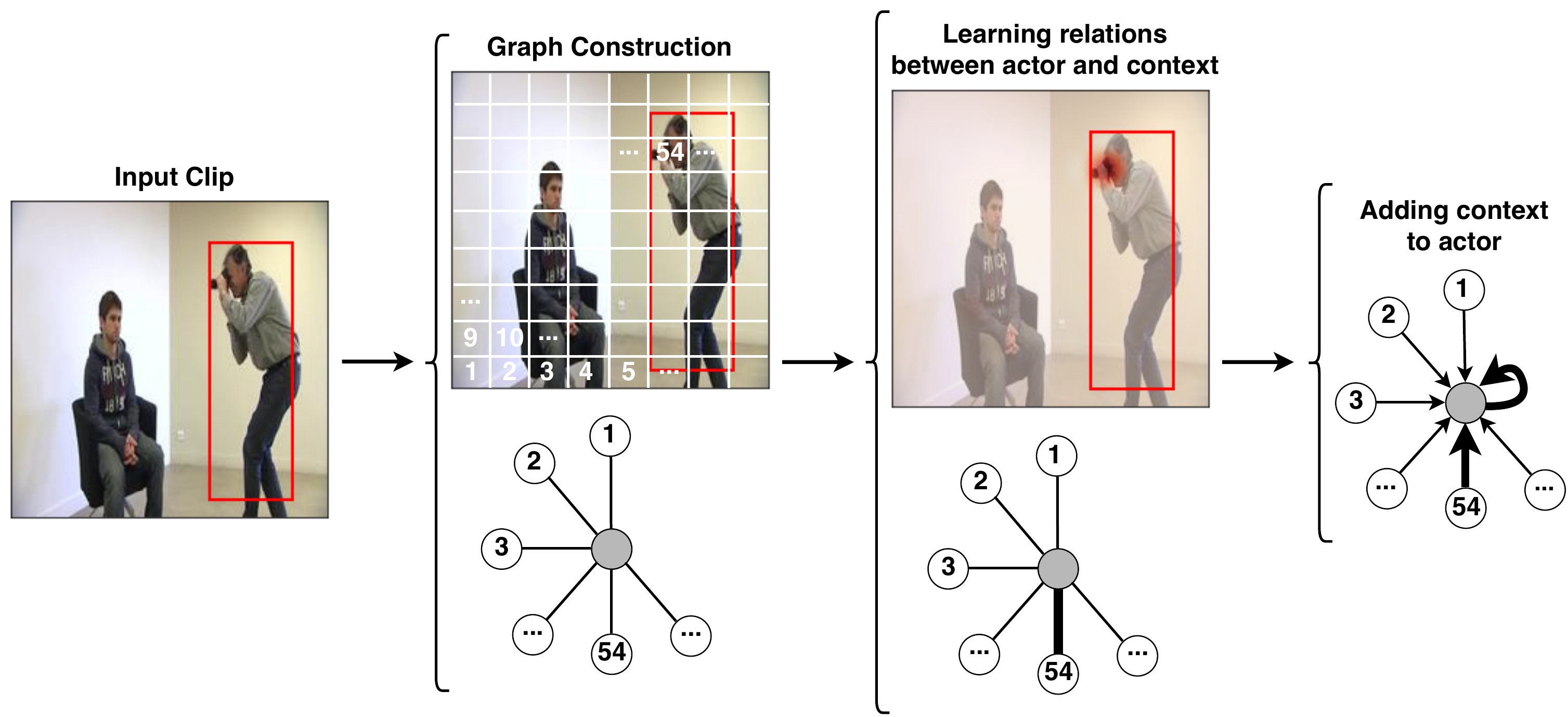}
    \caption{Given spatiotemporal features extracted from an input clip, we construct a graph with an actor (grey) node and context (numbered) nodes in order to model relations, such as actor-actor and actor-object interactions. Graph convolutions accumulate the learned context to the actor to obtain updated actor features for classification.}
    \label{taking_photos}
\end{figure}

In contrast to object detection, action detection requires learning of both appearance and motion features. Although spatiotemporal features are essential for action recognition and detection, they might prove insufficient for actions that share similar characteristics in terms of appearance and motion. For example, spatiotemporal features might not be sufficient to differentiate the action "Taking Photos" in Fig. \ref{taking_photos} from a similar one, such as "Phoning", since both share similar  characteristics in space and time (i.e. similar posture, motion around the head). As humans, we make use of context to put actions and objects in perspective, which can be an important cue to improve action recognition. Such contextual cues can refer to actor-object and actor-actor interactions.  For instance, a person holding a camera is more likely to perform the action "Taking Photos" than "Phoning", and vice versa. CNNs are able to capture such abstract or distant visual interactions only implicitly by stacking several convolutional layers, which increases the overall complexity and number of parameters. Hence, an approach to explicitly model contextual cues would be beneficial.

We introduce an approach to explicitly learn contextual cues, such as actor-actor and actor-object interactions, to aid action classification for the task of action detection. Our model, inspired by recent work on graph neural networks \cite{Kipf:2016tc,Velickovic2018GraphAN,DBLP:journals/corr/abs-1904-10117}, learns context by performing relational reasoning on a graph structure using Graph Convolutional Networks (GCN) \cite{Kipf:2016tc}. A high-level overview is illustrated in Fig.~\ref{taking_photos}. Given a detected actor in a short video clip, we construct a graph with an actor node encoding actor features, and context nodes encoding context features, such as objects and other actors in the scene. The graph's adjacency matrix consists of relation values encoding the importance of context nodes to the actor node, and is learned during training via gradient descent. Graph convolutions accumulate the learned context to the actor to obtain contextualized/updated actor features for action classification. Our model aids explainability by visualizing the learned adjacency matrix as an attention map that highlight the relevant context for recognizing the action.

We are interested in an approach to learn these contextual cues using as few annotated data as possible. Recent works \cite{Girdhar2019VideoAT,Sun2018ActorCentricRN,Ulutan2020ActorCA,8953450} that model contextual cues for action detection rely on full supervision in terms of actor bounding box annotations. However, extensive video annotation is time consuming and expensive \cite{10.1007/s11263-018-1120-4}. In this work, we are interested in learning context for the task of weakly-supervised action detection, i.e. action detection when only a handful of annotated frames are available throughout the action instance. Following the setting of sparse spatial supervision \cite{DBLP:journals/corr/WeinzaepfelMS16}, we train our contextual model by using up to five actor bounding box annotations throughout the action instance. 

We evaluate our models on the challenging Daily Action Localization in YouTube (DALY) dataset \cite{DBLP:journals/corr/WeinzaepfelMS16}, which consists of 10 action classes of human-object interactions (e.g. Drinking, Phoning, Brushing Teeth), and is annotated based on sparse spatial supervision. Therefore, DALY is a suitable test bed to model context for the task of weakly-supervised action detection.

Our contributions are as follows: 1) We introduce an architecture employing Graph Convolutional Networks \cite{Kipf:2016tc} in order to model contextual cues to improve classification of human actions in videos; 2) Our model aids explainability by visualizing the graph's adjacency matrix in the form of attention maps that highlight the learned context, even in a zero-shot setting, i.e. for actions and objects unseen during training; 3) We achieve 1) and 2) in a weakly-supervised setting, i.e. when annotated data are sparse throughout the action instance; 4) We introduce an intuitive metric based on recall of retrieved objects in attention maps, in order to quantitatively evaluate how well the model highlights the important context. As attention maps are often used for qualitative inspection only, this metric may be of general use beyond our use case.

In Section \ref{methods_subsection}, we present the baseline model and our approach on learning context by performing reasoning on a graph structure using GCN \cite{Kipf:2016tc}. Additionally, we discuss training of these models using sparse spatial supervision \cite{DBLP:journals/corr/WeinzaepfelMS16}. We conduct experiments and report results in Section \ref{experimental_setup_section}. In Section \ref{analysis_section}, we perform a qualitative and quantitative analysis of attention maps.

\section{Related Work}\label{section_related_work}

\subsection{Action Recognition and Detection}

Action recognition aims to classify the action taking place in a video. Early approaches relied on two-stream 2D CNNs \cite{10.5555/2968826.2968890} operating on RGB and optical flow inputs, or on detectors to classify bounding boxes of components in keyframes leaving action classification at the video level to traditional machine learning techniques \cite{vanBovenetal2018}. Recent works focus on (two-stream) 3D CNNs \cite{8099985,8237852,8578773}, which perform spatiotemporal (3D) convolutions. While action recognition considers the classification task, action detection carries out both classification and detection of actions. Although action detection is usually addressed using full supervision \cite{Girdhar2019VideoAT,Gkioxari2015FindingAT,Sun2018ActorCentricRN,Ulutan2020ActorCA,7410719,8953450}, we are interested in weak supervision, which allows us to reduce annotation cost by training models using very few action bounding box annotations per action instance.

Sivan and Xiang \cite{BMVC.25.65} approach weakly-supervised action detection using Multiple Instance Learning (MIL). Their approach requires binary labels at the video level, indicating the presence of an action.  Mettes et al. \cite{10.1007/s11263-018-1120-4} propose action annotation using points, instead of boxes. Chéron et al. \cite{NIPS2018_7373} present a unified framework for action detection by incorporating varying levels of supervision in the form of labels at the video level, a few bounding boxes, etc. Weinzaepfel et al. \cite{DBLP:journals/corr/WeinzaepfelMS16} introduce DALY and the setting of sparse spatial supervision, in which, up to five bounding boxes are available per action instance. Chesneau et al. \cite{DBLP:journals/corr/ChesneauRAS17} produce full-body actor tubes inferred from detected body parts, even when the actor is occluded or part of the actor is not included in the frame. 

\subsection{Visual Relational Reasoning}
There has been recent research on augmenting deep learning models with the ability to perform visual relational reasoning. 

Santoro et al. \cite{Santoro2017ASN} propose the relation network, which models relations between pairs of feature map pixels for visual question answering. This idea has been extended for action detection \cite{Sun2018ActorCentricRN,Ulutan2020ActorCA} to model actor-context relations. Similar to \cite{Sun2018ActorCentricRN,Ulutan2020ActorCA}, we treat every 1$\times$1$\times$1 location of the feature map as context. In these works, learned actor-context relations are used either directly \cite{Sun2018ActorCentricRN} or to highlight features of the feature map \cite{Ulutan2020ActorCA} for action classification. In contrast, we encode actor-context relations as edges in a graph, and graph convolutions output updated actor features for action classification. 

With regard to visual attention, non-local neural networks \cite{8578911} compute the output of a feature map pixel as a weighted sum of all input pixels. For action detection, Girdhar et al. \cite{Girdhar2019VideoAT} extend the Transformer architecture \cite{DBLP:journals/corr/VaswaniSPUJGKP17} for action detection. Although a Transformer can be represented as a graph neural network and vice versa, we argue that a graph representation is simpler and more intuitive compared to the Transformer representation of Queries, Keys and Values. Whilst \cite{Girdhar2019VideoAT} use two transformations of a feature map to represent context as Keys and Values, this representation does not have a direct interpretation in a graph. In contrast, we use a single transformation to obtain context features, which correspond to context nodes in the graph. Furthermore, our model does not require residual connections \cite{7780459} nor Layer Normalization \cite{ba2016layer}, two essential components of the Transformer. 

In this work, we apply Graph Convolutional Networks (GCN) \cite{Kipf:2016tc}, which provide a structured and intuitive way to model relations between nodes in a graph. Recently, GCN have been used for visual relational reasoning for the tasks of action recognition \cite{10.1007/978-3-030-01228-1_25} and group activity recognition \cite{DBLP:journals/corr/abs-1904-10117}. Zhang et al. \cite{8953450} employ GCN \cite{Kipf:2016tc} for action detection, where nodes represent detected actors and objects. As we do not require an external object detector, our approach is suitable to reason with respect to arbitrary context and objects which cannot be detected, e.g. because the object detector has not been trained to do so. 

Whilst aforementioned approaches \cite{Girdhar2019VideoAT,Sun2018ActorCentricRN,Ulutan2020ActorCA,8953450} rely on full actor supervision during training, our work focuses on learning contextual cues using weak supervision in terms of actor bounding box annotations. Although similar attention maps are also presented in \cite{Girdhar2019VideoAT,Sun2018ActorCentricRN,Ulutan2020ActorCA}, our work is the first to generate them in a zero-shot setting, and introduce a metric to quantitatively evaluate them. 

\section{Learning Context with GCN} \label{methods_subsection}

\begin{figure}[tb]
    \centering
    \includegraphics[width=.9\textwidth]{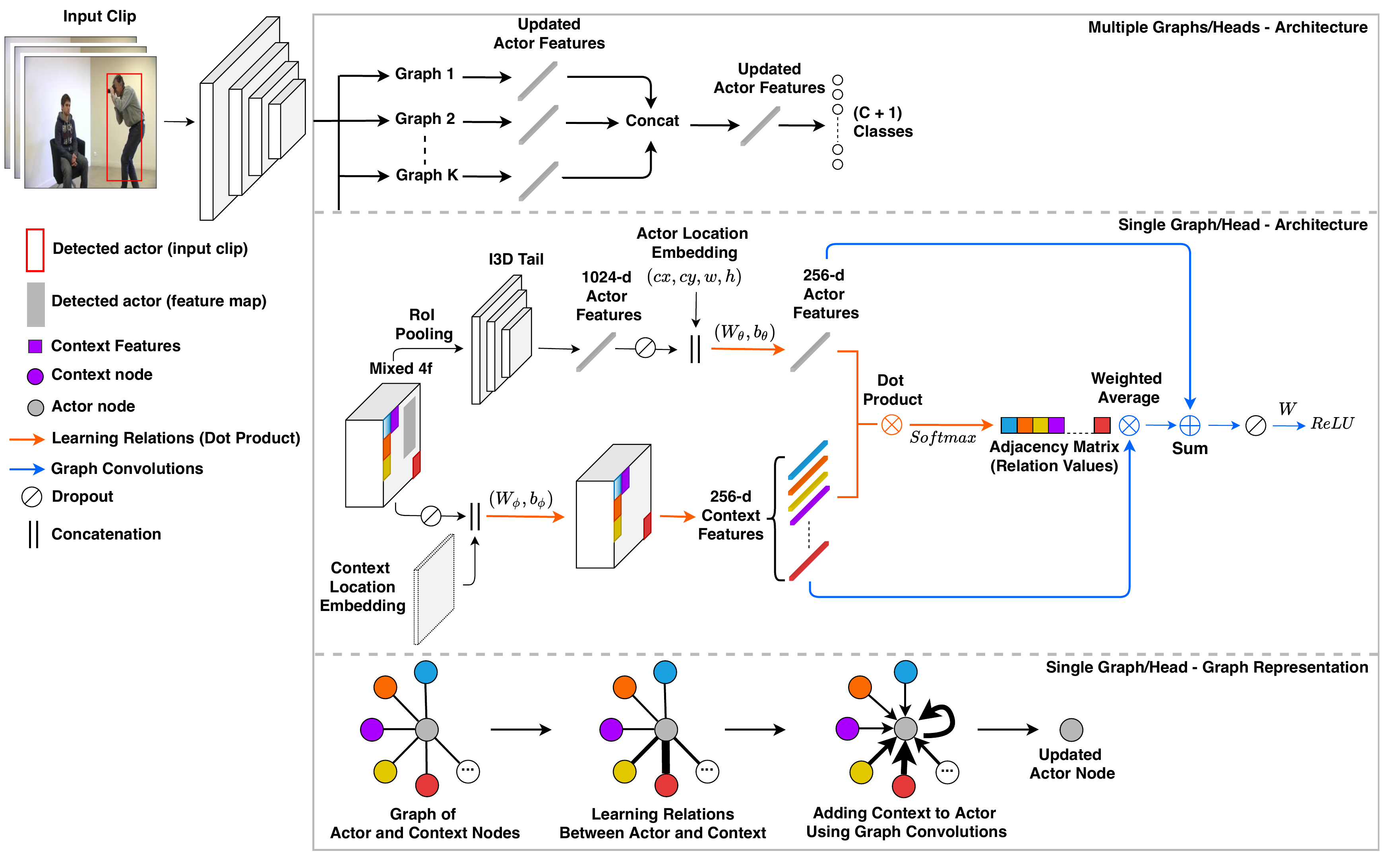}
    \caption{The lowest part shows the graph representation of the GCN model for a single graph, while its implementation using matrix operations is shown in the middle part. The top part illustrates the construction of multiple graphs (multi-head attention) in order to learn different types of actor-context relations.}
    \label{framework_overview}
\end{figure}

We propose an approach to learn contextual cues, such as actor-actor and actor-object interactions, by performing relational reasoning on a graph structure using Graph Convolutional Networks (GCN) \cite{Kipf:2016tc}. We expect such contextual cues to improve action classification, as they can be discriminative for the actions performed, and provide more insight into what the model has learnt, which benefits interpretability. Moreover, we are interested in learning context using weak supervision in terms of actor bounding box annotations. An overview of our proposed GCN model is illustrated in the second branch of Fig. \ref{framework_overview}. The input is a short video clip with at least one actor performing an action. A 3D convolutional network extracts spatiotemporal features for the input clip, up to a convolutional layer. We treat every $1\times1\times1$ spatiotemporal location of the output feature map as context, while actor features are extracted by RoI Pooling \cite{7410526} on the detected actor's bounding box and subsequent 3D convolutional layers. We construct a graph consisting of context nodes and an actor node, with connections drawn from every context node to the actor node. Relation values between nodes are encoded in the adjacency matrix, and learned using a dot-product self-attention mechanism \cite{DBLP:journals/corr/VaswaniSPUJGKP17,Velickovic2018GraphAN}. Graph convolutions accumulate the learned context to the actor in order to obtain updated/contextualized actor features for action classification. We compare the GCN model to a baseline model that uses no context, and classifies the action using the feature representation corresponding to the actor bounding box. 


In this section, we first present the 3D convolutional backbone network to learn spatiotemporal features using weak supervision. Next, we present our approach on learning contextual cues for action detection using GCN and we provide implementation details.  

\subsection{Feature Extraction with Weak Supervision} \label{feature_extraction}


\subsubsection{Backbone Network}

Spatiotemporal features for the whole input clip are extracted using a 3D convolutional backbone network. There are several 3D architectures in literature \cite{8099985,8237852,8578773}. We opt for I3D \cite{8099985}, which is widely used and has demonstrated very positive results in action recognition. The input is a sequence of frames of size $C\times T\times H\times W$, where $C$ denotes the number of channels,  $T$ is the number of input frames, and $H$ and $W$ represent the height and width of the input sequence. Features are extracted up to \texttt{Mixed\_4f} layer, which has an output feature map of size $D'\times T'\times H' \times W'$, where $D'$ denotes the number of feature channels, $T' = \frac{T}{8}$, $H' = \frac{H}{16}$, and $W' = \frac{W}{16}$. 

\subsubsection{Actor Feature Extraction with Weak Supervision} 

We are interested in learning spatiotemporal features using only a handful of annotated frames per action instance. For an annotated frame, also called keyframe, annotation is in the form of an action bounding box and corresponding class label. Due to the limited number of available annotated frames, training a Region Proposal Network (RPN) \cite{7485869} to produce actor box proposals would be sub-optimal. To this end, we train our models using sparse spatial supervision as introduced in \cite{DBLP:journals/corr/WeinzaepfelMS16}. In detail, a Faster R-CNN \cite{7485869} detects all actors in each frame, and detections are tracked throughout the action instance using a tracking-by-detection approach \cite{7410719}, which produces class-agnostic action tubes. In practice, we use tubes provided by \cite{DBLP:journals/corr/WeinzaepfelMS16}. Tubes are labeled based on spatiotemporal Intersection over Union (IoU) with sparse annotations, i.e. ground truth tubes comprised of up to 5 bounding boxes throughout the action instance. Tubes with spatiotemporal IoU greater than 0.5 are assigned to the action class of the ground truth tube with the highest IoU. If no such ground truth tube exists, the action tube is labeled as background. The backbone is augmented with a RoI pooling layer \cite{7410526} to extract features for each actor for action classification. Boxes of each tube are appropriately scaled and mapped to the output feature map of \texttt{Mixed\_4f} layer, with a temporal stride of four frames. For each action tube, RoI pooling extracts actor features of size $D' \times T' \times 7 \times 7$. Actor features are then passed through I3D tail consisted of 3D convolutional layers \texttt{Mixed\_5b} and \texttt{Mixed\_5c}. Finally, a spatiotemporal (3D) average pooling layer reduces the size to $D'' \times 1 \times 1 \times 1$.

\subsection{Graph Convolutional Networks} \label{graph_convolutional_networks}

\subsubsection{Learning Relations}

Our graph consists of two types of nodes: context nodes and actor nodes. Context node features, $f_j' \in \mathbb{R}^{D'\times1\times1\times1}$, $j=1,2,\dots, M$, $M = T'H'W'$, correspond to every $1\times 1\times 1$ spatiotemporal location of the output feature map of \texttt{Mixed\_4f} layer. Actor node features, $a_i' \in \mathbb{R}^{D''\times 1}$, $i=1,2,\dots,N$, where $N$ is the number of detected actors in the input clip, are extracted as described in Section \ref{feature_extraction}. Relations between actor features and context features, shown in orange arrows in Fig. \ref{framework_overview}, are learned using a dot-product self-attention operation \cite{DBLP:journals/corr/VaswaniSPUJGKP17,Velickovic2018GraphAN}, after projecting the features in a lower dimensional space using a linear transformation. Formally, 
\begin{align}
    e_{ij} = \theta(a_i')^T \cdot \phi(f_j')
\end{align}
where
\begin{align}
    a_i = \theta(a_i') = \mathbf{W}_{\theta}a_i' + \mathbf{b}_{\theta}\label{actor_transformation}\\ 
    f_j = \phi(f_j') = \mathbf{W}_{\phi}f_j' + \mathbf{b}_{\phi}\label{context_transformation}
\end{align}
Eq. \ref{actor_transformation}--\ref{context_transformation} are transformations for actor features and context features, respectively, with $\mathbf{W}_{\theta} \in \mathbb{R}^{D''\times D}, \mathbf{W}_{\phi} \in \mathbb{R}^{D'\times D}$; $\mathbf{b}_{\theta}, \mathbf{b}_{\phi} \in \mathbb{R}^{D\times 1}$; $D < D',D''$. In matrix form, $\mathbf{A} \in \mathbb{R}^{N\times D}$ for transformed actor features and $\mathbf{F} \in \mathbb{R}^{M\times D}$ for transformed context features. The graph is represented by an adjacency matrix, $\mathbf{G} \in \mathbb{R}^{N\times M}$, where $g_{ij} \in \mathbf{G}$ denotes the relation or attention value, indicating the importance of context feature, $f_j$, to actor feature, $a_i$. Consequently, $\mathbf{G}$ is a directed graph connecting every context node to every actor node. Relation or attention values, $g_{ij}$, are obtained by applying softmax normalization on $e_{ij}$ (output of dot-product) across context features
\begin{align}
    g_{ij} = \frac{\exp(e_{ij})}{\sum_k \exp(e_{ik})}
\end{align}

\subsubsection{Graph Convolutions} \label{graph_convolutions_subsection}
Having defined the graph and a mechanism for learning actor-context relations, we perform reasoning on the graph in order to obtain updated actor features. This is achieved by accumulating information from context nodes to the actor node using graph convolutions. Updated actor features, $\mathbf{Z} \in \mathbb{R}^{N\times D}$, are obtained by
\begin{align}\label{gcn_layer}
    \mathbf{Z} = \sigma\Big(\Big(\mathbf{G} \mathbf{F} + \mathbf{A}\Big) \mathbf{W}\Big)
\end{align}
The operation is shown in blue arrows in Fig. \ref{framework_overview}. The weighted average of $\mathbf{F}$ with the relation values $\mathbf{G}$ produces weighted context features. Adding actor features $\mathbf{A}$ to the resulting representation imposes identity links for all actor nodes in the graph. The output is passed through a learnable linear transformation $\mathbf{W} \in \mathbb{R}^{D\times D}$ and a non-linear activation function $\sigma(\cdot)$ implemented as ReLU \cite{7410480}. 

In order to capture multiple types of relations between the actor and the context, we perform multi-head attention \cite{DBLP:journals/corr/VaswaniSPUJGKP17} by constructing multiple graphs at a given layer and merging their outputs using concatenation or summation. Weight matrices $\mathbf{W}_{\theta}, \mathbf{W}_{\phi}$, $\mathbf{W}$ are independent across graphs. Finally, in order to encode updated actor features on a higher level, we stack multiple GCN layers by providing the output of multiple graphs as input to the next GCN layer.  

\subsubsection{Location Embedding} \label{location_embedding}

Location information, such as the position of an actor with respect to other actors and objects, is important for modeling contextual cues. However, such information, encoded indirectly by regular convolutions, is lost when applying convolutions on a graph structure. 

We incorporate location information in both context features and actor features. For context features, we concatenate coordinates $(x, y)$ along the channel dimension before applying $\mathbf{W}_{\phi}$, indicating the location of the feature on the output feature map. For actor features, we concatenate coordinates $(cx, cy, w, h)$ before applying $\mathbf{W}_{\theta}$, corresponding to the average center, width and height of the actor tube across the input clip. Coordinates are normalized in $[-1, 1]$. 

\subsection{Implementation Details} \label{implementation_details_subsection}
 
We implement our models in PyTorch \cite{NEURIPS2019_9015}. I3D is pre-trained on ImageNet \cite{10.1007/s11263-015-0816-y} and then on the Kinetics \cite{8099985} action recognition dataset, while the external detector is pre-trained on the MPII Human Pose dataset \cite{6909866}. The input is a clip of 32 RGB frames with spatial resolution of $224\times224$. The output feature map of \texttt{Mixed\_4f} layer has $D'=832$ channels, while actor features have $D''=1024$. Transformations $\mathbf{W}_{\theta}$, $\mathbf{W}$ are implemented as fully connected layers and $\mathbf{W}_{\phi}$ as a 3D convolutional layer with kernel size $1\times 1\times 1$. We set $D=256$. We apply 3-dimensional dropout to context features before $\mathbf{W}_{\phi}$. Additionally, 1-dimensional dropout is applied to actor features before $\mathbf{W}_{\theta}$ in the first GCN layer, before $\mathbf{W}$ in all GCN layers and prior to the final classification layer (in both GCN and baseline model). Dropout probability is 0.5 in all cases. All fully connected layers are initialized using a Normal distribution according to \cite{7410480}. We set the gain parameter to 1 for $\mathbf{W}_{\theta}$ and to $\sqrt{2}$ for the rest of the fully connected layers. $\mathbf{W}_{\phi}$ is initialized using a Uniform distribution according to \cite{7410480} in the range $(-b + 0.01, b - 0.01)$ for the first GCN layer, and in the range $(-b, b)$ for subsequent layers, using a gain of $\frac{1}{\sqrt{3}}$. Biases of all layers are initialized to zero.

Models are optimized using SGD and cosine learning rate annealing, with learning rate $2.5\cdot10^{-4}$ over 150 epochs, and $4.7\cdot10^{-5}$ over 450 epochs, for the baseline and GCN model, respectively. We use a batch size of 3 clips, where each clip is randomly sampled from a video in the training set. Tubes of each clip are scored using the softmax scores produced by the model. During inference, we sample 10 32-frame clips from each video, and tubes are scored by averaging the softmax scores across the clips. The same clips are sampled in order to facilitate fair comparison between different models. Training time is approximately one day for GCN and less than half a day for the baseline on a GTX 1080 Ti GPU. 

\section{Experiments} \label{experimental_setup_section}
In this section, we first describe the DALY dataset and the evaluation metric used throughout the experiments. Next, we conduct experiments to evaluate the performance of the GCN model, and we compare it with the baseline and the state of the art on DALY. Finally, we evaluate the GCN model using minimal spatial supervision i.e. one bounding box per action instance.

\subsection{Dataset and Evaluation Metric} \label{dataset_section}

We develop and evaluate our models on the Daily Action Localization in Youtube (DALY) \cite{DBLP:journals/corr/WeinzaepfelMS16} dataset. It consists of 510 videos of 10 human actions, such as "Drinking", "Phoning" and "Brushing Teeth". In this paper, we do not perform temporal localization, and we assume that the temporal boundaries of each action instance within a video are known. An action instance has an average duration of 8 seconds and may contain more than one person performing an action. Each of the 10 classes contains an interaction between a person and an object that define the action taking place. There are 31 training videos and 20 test videos per class. We fine-tune our models by holding out a subset of the training set as a validation set, consisted of 10 videos from each class. We evaluate models using Video-mAP at 0.5 IoU threshold (Video-mAP@$0.5$) \cite{Gkioxari2015FindingAT}. 

\subsection{Evaluation of Architecture Choices} \label{num_layers_and_graphs_subsection_results}
In this section, we experimentally evaluate the GCN model with respect to several architecture choices. Specifically, we experiment with up to two GCN layers and up to three graphs per layer. Additionally, we compare concatenation and summation as merging functions to combine the output of multiple graphs. Finally, we measure the impact of including the location embedding and the I3D tail (convolutional layers \texttt{Mixed\_5b} and \texttt{Mixed\_5c}) to extract actor features. 

Results with respect to different number of layers and graphs are shown in Table \ref{num_layers_num_graphs_results}, along with the number of parameters for every configuration (I3D parameters are not included). Note that for two GCN layers, the first layer always employs concatenation as a merging function. Building multiple graphs is beneficial for model performance, for both functions. It is interesting that mAP increases for a 2-layer GCN model with concatenation, but not with summation. Concatenation outperforms summation in nearly all configurations. For the rest of the experiments, we choose a 2-layer, 2-graph GCN model with concatenation, which provides a good trade-off between performance and number of parameters.

In order to measure the impact on model performance obtained by the location embedding and I3D tail, we remove them from the architecture and examine the difference in model performance. By removing the location embedding, the model has no information of the actor's location relatively to other actors and objects, and relations are calculated based solely on visual features. This results in a decrease of 1.1 points in mAP (50.7), which indicates that modeling spatial actor-context relations improves performance. By removing the I3D tail, actor features are extracted from the output feature map of \texttt{Mixed\_4f} layer. This results in a significant decrease of more than four points in mAP (47.42), highlighting the importance of using the I3D tail to encode actor features.

\begin{table}[tb]
    \centering
    \caption{Validation mAP with respect to different number of layers, number of graphs per layer and merging functions to combine the output of multiple graphs. The number of model parameters are provided for every configuration.}
    \label{num_layers_num_graphs_results}
    
    \begin{tabular}{c|c|c|c|c}
        \toprule
        \# Layers& \# Graphs & Merging Function & \# Parameters & Val. mAP\\
        \midrule
        1 & 1 & - & 543K & 49.39\\
        \midrule
        \multirow{2}{*}{1} & \multirow{2}{*}{2} & Sum & 1.084M & 51.39\\
         & & Concat & 1.086M & 51.3\\
        \midrule
        \multirow{2}{*}{1} & \multirow{2}{*}{3} & Sum & 1.624M & 50.7\\
         &  & Concat & 1.630M & 50.98\\
        \midrule
        2 & 1 & - & 887K & 47.28\\
        \midrule
        \multirow{2}{*}{2} & \multirow{2}{*}{2} & Sum & 1.903M & 50.1\\
         &  & Concat & 1.906M & 51.82\\
        \midrule
        \multirow{2}{*}{2} & \multirow{2}{*}{3} & Sum & 3.050M &49.98\\
         &  & Concat & 3.055M & 52.09\\
        \bottomrule
    \end{tabular}
\end{table}

\subsection{Comparison with Baseline and State of the Art} \label{comparison_sota_subsection_results}
We compare the GCN model with the baseline model and the state-of-the-art \cite{DBLP:journals/corr/ChesneauRAS17,DBLP:journals/corr/WeinzaepfelMS16} on the DALY test set in Table \ref{comparison_results_table}.

\begin{figure}[tb]
    \centering
    \begin{minipage}{.62\textwidth}
        \centering
        \includegraphics[width=\textwidth]{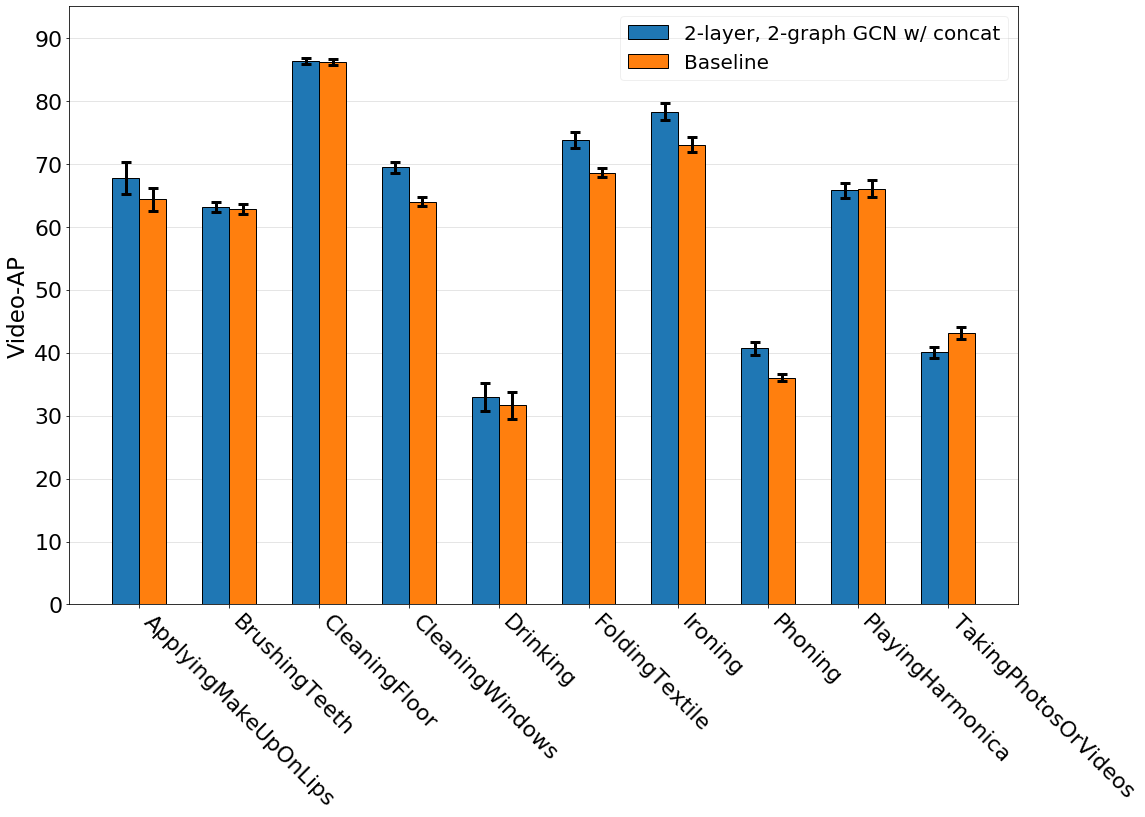}
    \end{minipage}%
    \begin{minipage}{.38\textwidth}
        \centering
        \includegraphics[width=\textwidth]{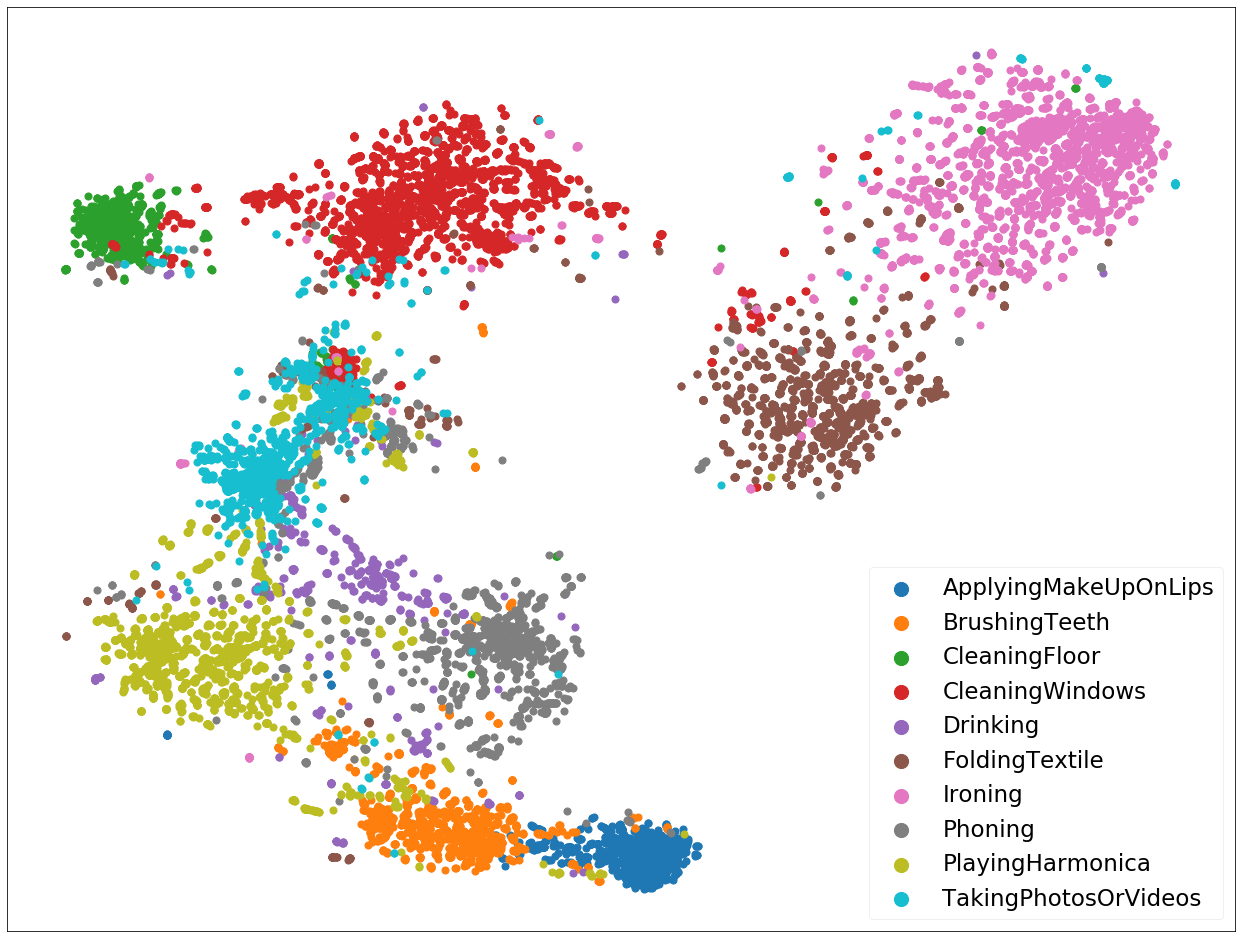}
        \includegraphics[width=\textwidth]{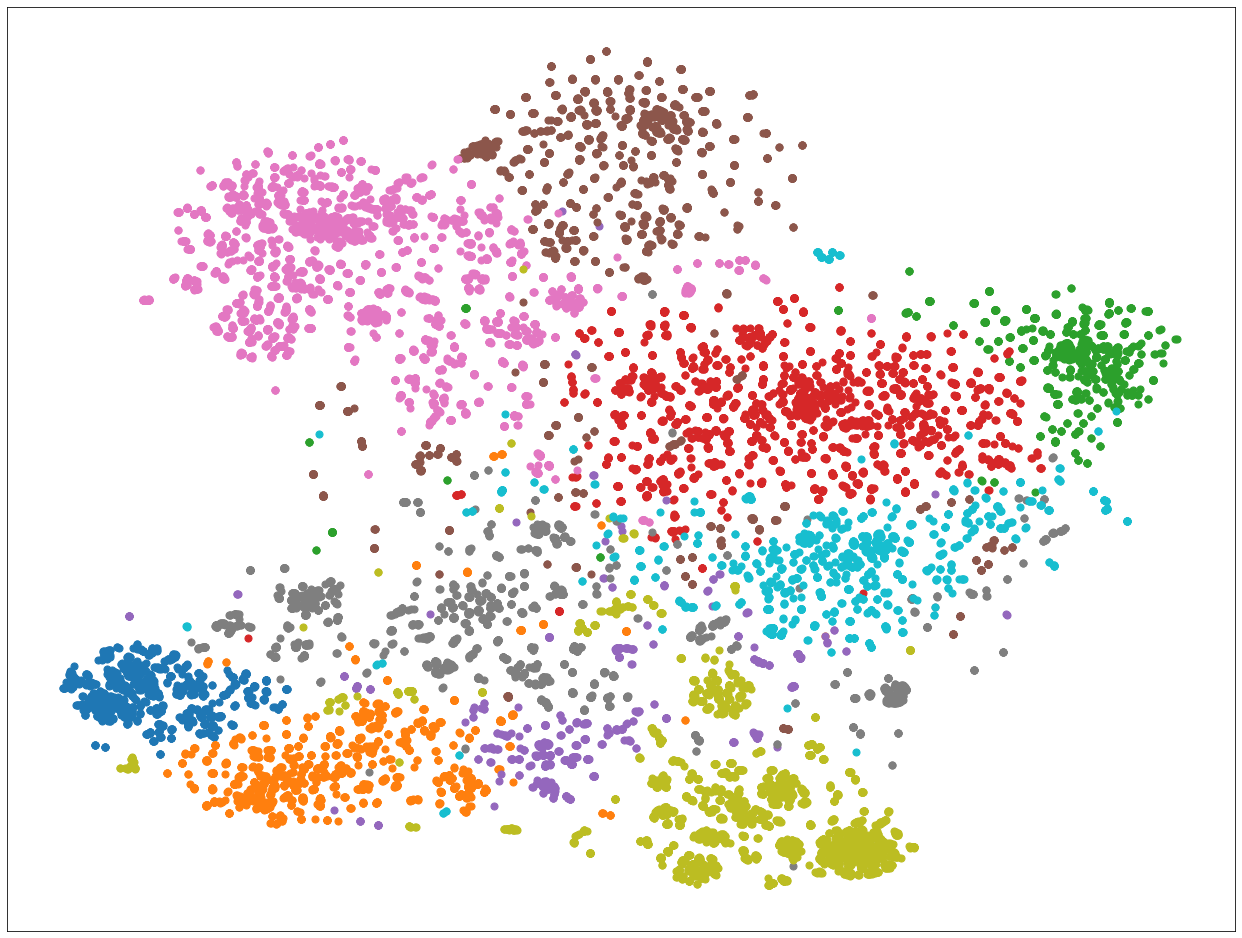}
    \end{minipage}
    \caption{Per-class Video-AP on the test set across five repetitions of GCN
and baseline model, and t-SNE actor feature embeddings of GCN (top) and
baseline model (bottom).}
    \label{barplot_ap}
\end{figure}

The baseline model classifies actor features obtained from I3D (see Section \ref{feature_extraction}) using a linear layer that outputs classification scores for $C$ action classes and a background class. Results are shown in Table \ref{comparison_results_table}. Across five repetitions, the GCN model outperforms the baseline model by 2.24 (3.7\%) points in mean mAP, and by 2.94 (4.9\%) points in maximum mAP. The left-hand side of Fig. \ref{barplot_ap} illustrates per-class average precision for the baseline and GCN model. GCN performs comparably or better than the baseline model in all classes except "TakingPhotosOrVideos". On the right-hand side of Fig. \ref{barplot_ap}, we visualize t-sne \cite{vanDerMaaten2008} actor feature embeddings, colored by the respective action, for the GCN (top) and baseline model (bottom). The GCN model produces tighter and more distinct clusters compared to the baseline model. 

Comparing the GCN model with the state-of-the-art \cite{DBLP:journals/corr/ChesneauRAS17,DBLP:journals/corr/WeinzaepfelMS16}, we obtain slightly improved performance in comparison to \cite{DBLP:journals/corr/WeinzaepfelMS16}, while Chesneau et al. \cite{DBLP:journals/corr/ChesneauRAS17} achieve better performance by 1.69 points in mean mAP and 0.78 points in maximum mAP. We argue that this is due to the following reasons. Firstly, our models are trained using fewer videos, since we hold out a part of the training set as a validation set for fine-tuning. Secondly, \cite{DBLP:journals/corr/ChesneauRAS17,DBLP:journals/corr/WeinzaepfelMS16} train their models on the region proposals produced by the detector (see Section \ref{feature_extraction}), while we train our models on the data provided by \cite{DBLP:journals/corr/WeinzaepfelMS16}, which are only the final detections of the detector. Consequently, we train our models using fewer videos and boxes compared to \cite{DBLP:journals/corr/ChesneauRAS17,DBLP:journals/corr/WeinzaepfelMS16}. It is worth noting that, in contrast to \cite{DBLP:journals/corr/ChesneauRAS17,DBLP:journals/corr/WeinzaepfelMS16}, our model does not employ expensive optical flow computation.

\begin{table}[tb]
    \centering
    \caption{Comparison of GCN model with the baseline model and state-of-the-art on the test set. We report model architecture and input modalities (RGB, Optical Flow).}
    \begin{tabular}{c|c|c|c}
        \toprule
        Model & Architecture & Input & Test mAP \\
        \midrule
        Weinzaepfel et al. \cite{DBLP:journals/corr/WeinzaepfelMS16} & Fast R-CNN (VGG-16) & RGB, OF & 61.12\\
        Chesneau et al. \cite{DBLP:journals/corr/ChesneauRAS17} & Fast R-CNN (VGG-16) & RGB, OF & 63.51\\
        Baseline (Ours) & I3D & RGB & 59.58 ($\pm$ 0.22) (59.79)\\
        GCN (Ours) &  I3D & RGB & 61.82 ($\pm$ 0.51) (62.73)\\
        \bottomrule
    \end{tabular}
    \label{comparison_results_table}. 
\end{table}

\subsection{Reducing Annotation to One Bounding Box} \label{reduced_annotation_subsection_results}
We examine model performance when minimal spatial supervision is used, i.e. one bounding box per action instance. We label tubes based on spatial IoU with the ground truth box of a randomly selected keyframe for each action instance. A GCN model is then trained using the newly labeled tubes. Using only a single keyframe to label tubes, we obtain a small decrease in mAP, from 61.82 to 61.07. On the other hand, when one keyframe is used during evaluation too, performance decreases by 3.15 points in mAP. The reason for such a decrease is that mAP is not adequately estimated using only one evaluation keyframe.

\section{Analysis of Attention} \label{analysis_section}
Our GCN model aids explainability by visualizing the adjacency matrix in the form of attention maps that highlight the learned context, even in a zero-shot setting, i.e. for actions and objects unseen during training. Although similar attention maps are presented in previous works \cite{Girdhar2019VideoAT,Sun2018ActorCentricRN,Ulutan2020ActorCA} (albeit not in a zero-shot setting), in this paper, we go one step further to quantitatively evaluate the ability of the attention to highlight the relevant context. To this end, we propose a metric based on recall of objects retrieved by attention maps. In this Section, we present qualitative and quantitative results of attention maps. 

\subsection{Evaluation of Attention Maps} \label{attention_maps_subsection}

The adjacency matrix contains the relation or attention values, indicating the importance of every context node (spatiotemporal location of feature map) to the actor node. By visualizing the adjacency matrix, we obtain an attention map that highlights, for a given actor, the important context regions the model pays most attention to. The map is interpolated to the original input size and overlaid on the input clip. Our model is even able to generalize its attention in a zero-shot setting i.e. for actions that the model has not been trained to recognize. To achieve this, we train a GCN model by excluding two action classes, and then visualize the attention maps for the excluded classes.

\subsubsection{Qualitative Evaluation}

\begin{figure}[tb]
    \centering
    \begin{minipage}{.5\textwidth}
        \centering
            \includegraphics[width=\textwidth]{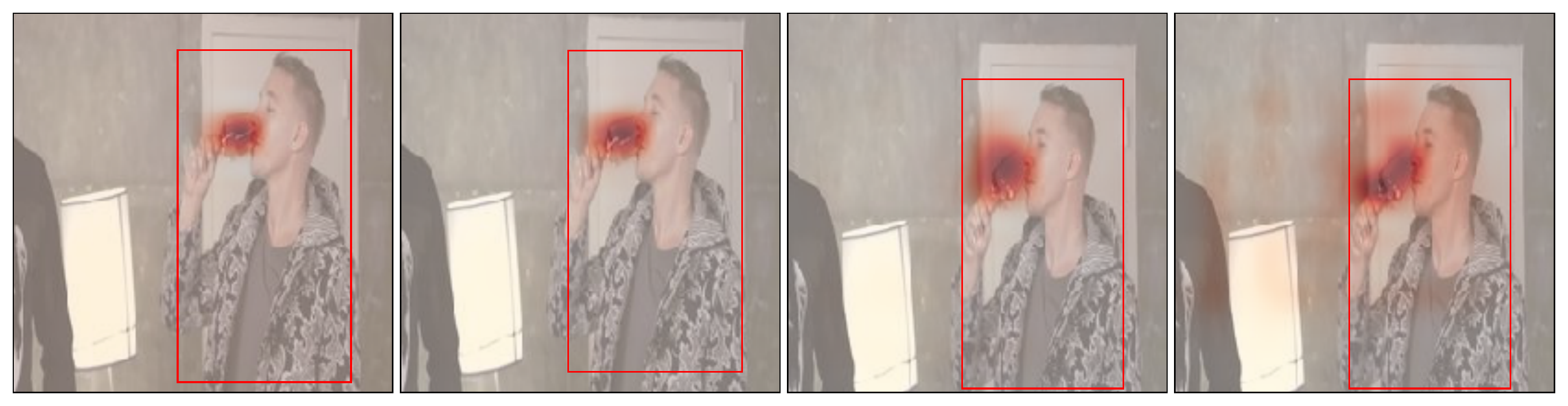}
            \includegraphics[width=\textwidth]{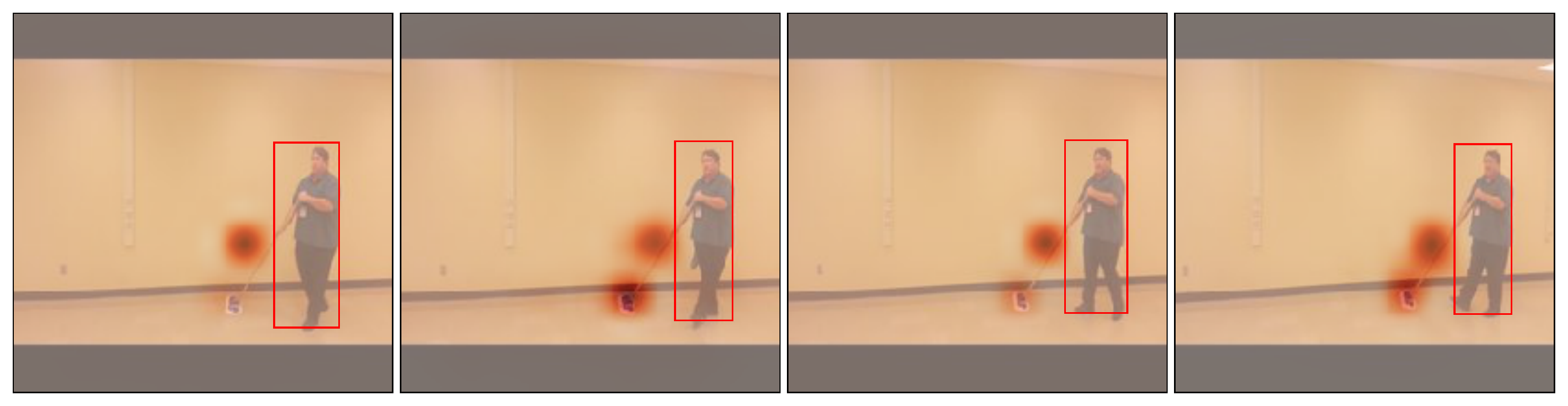}
            \includegraphics[width=\textwidth]{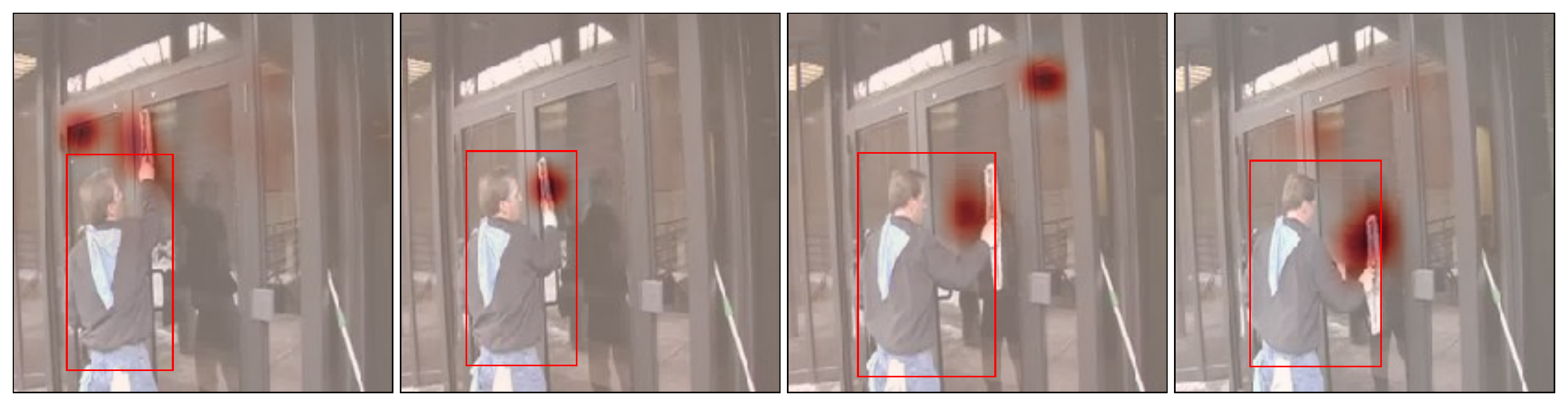}
    \end{minipage}%
    \begin{minipage}{.5\textwidth}
        \centering
        \includegraphics[width=\textwidth]{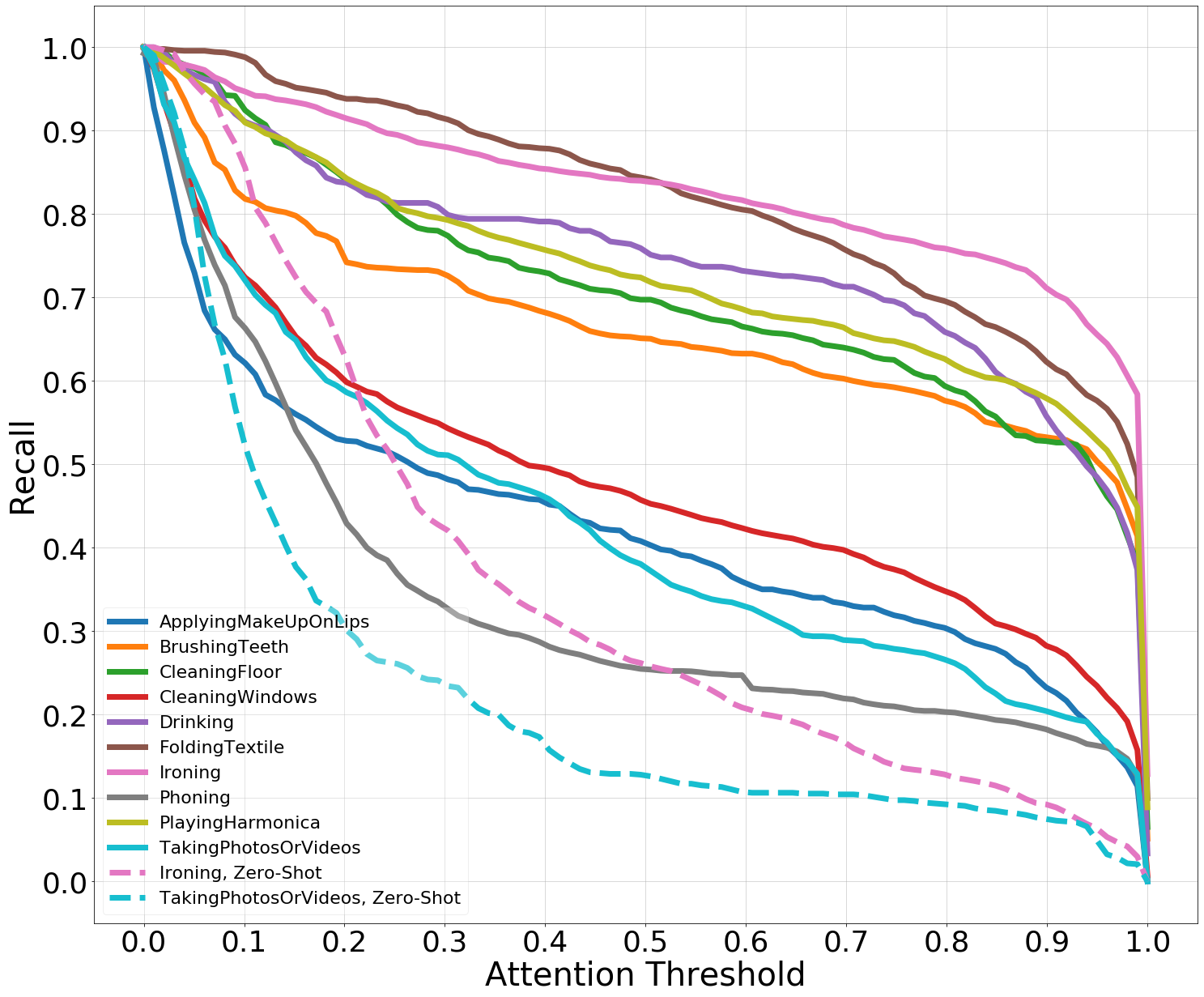}
    \end{minipage}
    \centering
    \includegraphics[width=.5\textwidth]{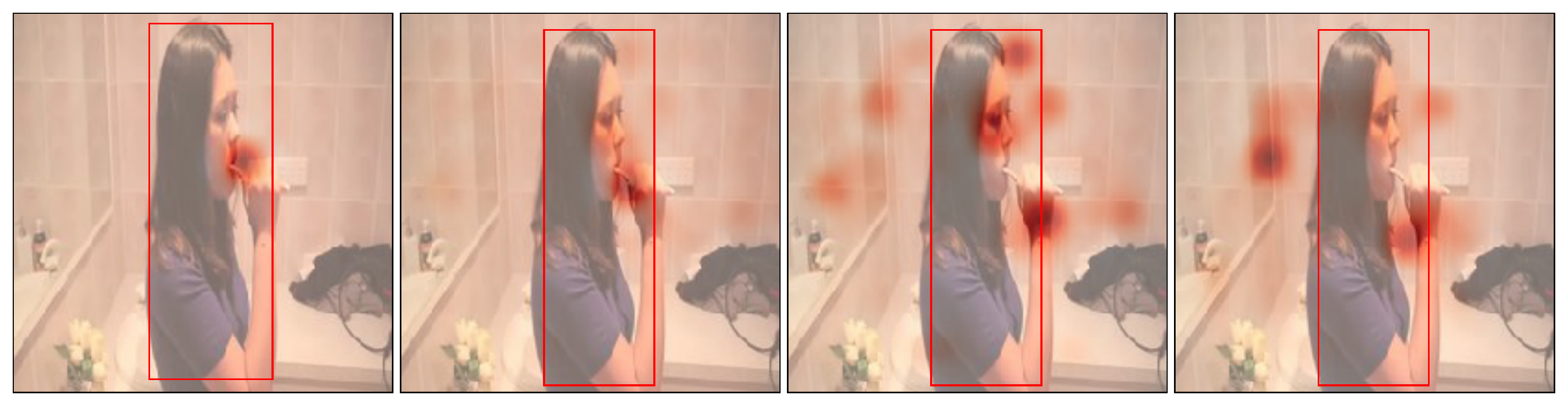}\includegraphics[width=.5\textwidth]{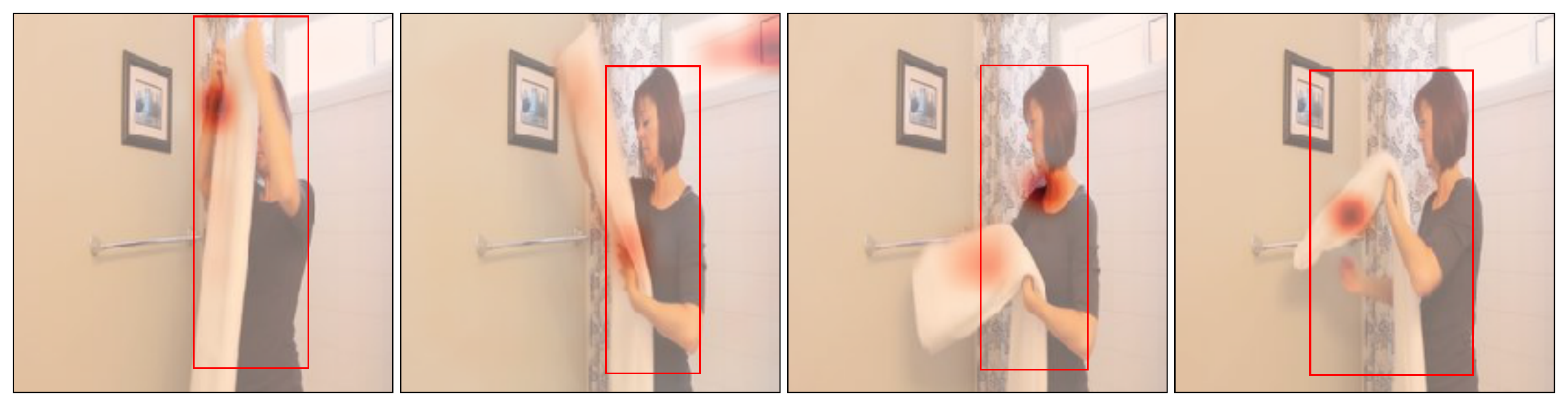}
    \includegraphics[width=.5\textwidth]{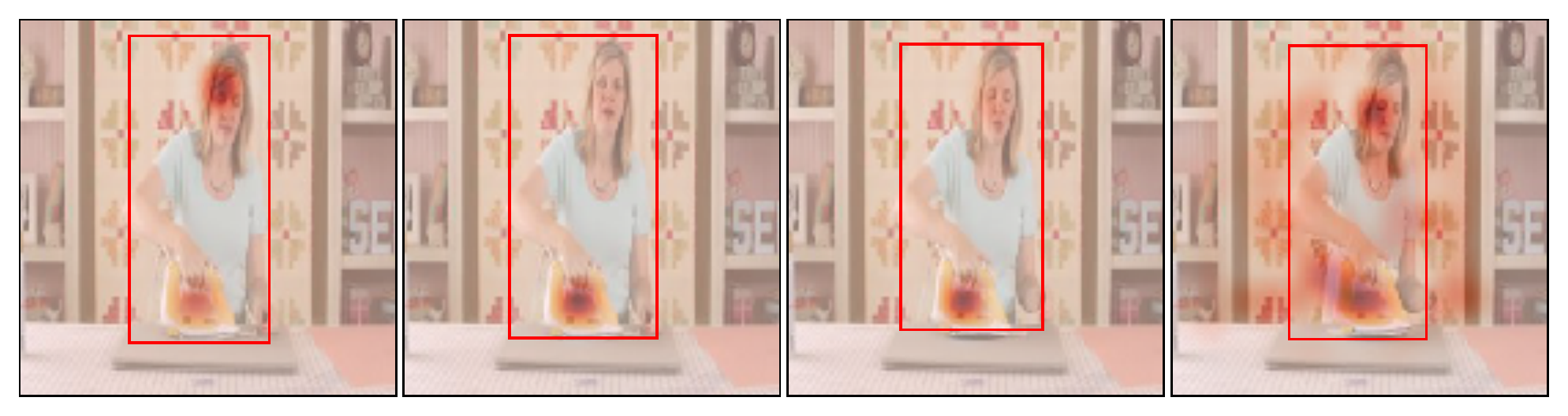}\includegraphics[width=.5\textwidth]{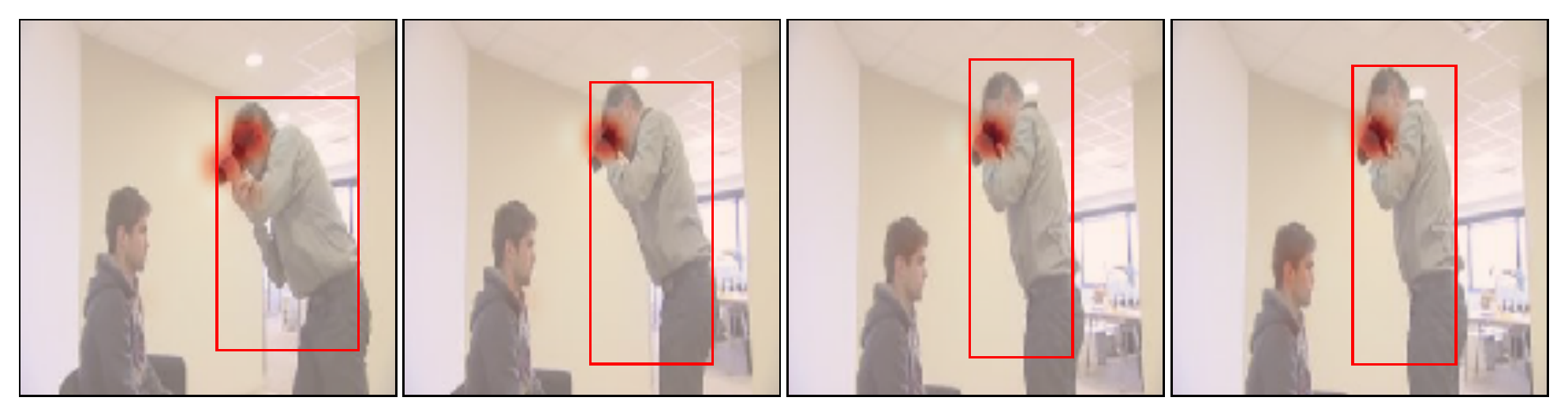}
    \caption{Per-class object recall curves, along with visualizations of attention maps (last row illustrates zero-shot cases) for actions ``Drinking", ``CleaningFloor", ``CleaningWindows", ``BrushingTeeth", ``FoldingTextile", ``Ironing", ``TakingPhotosOrVideos".}
    \label{attention_maps}
\end{figure}

Fig. \ref{attention_maps} illustrates examples of attention maps, where each one is the combination of four adjacency matrices (2 GCN layers; 2 graphs per layer) by summing their values along the spatial dimensions. Each example contains four attention maps, representing time progression along the input clip. The last row of Fig. \ref{attention_maps} contains zero-shot attention maps for classes "Ironing" and "TakingPhotosOrVideos". The attention maps show that our GCN model highlights relevant context, such as objects, hands and faces, and is also able to track objects along time. Finally, our model is able to highlight relevant objects (e.g. Iron, Camera) for actions unseen during training (last row of Fig. \ref{attention_maps}).

\subsubsection{Quantitative Evaluation}
We evaluate how well the attention maps highlight relevant objects by introducing a metric based on recall of objects retrieved by the attention. DALY provides object bounding box annotations on annotated frames. Given the attention map produced for a detected actor, we sum the attention values inside the object's bounding box. An instance is a true positive if the sum of values is larger than a threshold, and a false negative, otherwise. 

A per-class quantitative evaluation of attention maps is shown in Fig. \ref{attention_maps}, with recall on the $y$-axis and the attention threshold on the $x$-axis. Dashed curves correspond to zero-shot cases. Our metric suggests that objects are retrieved by the attention with relatively high recall, even for large attention thresholds, which shows the effectiveness of our model to highlight the relevant context.

\section{Conclusion} \label{conclusion_section}
We propose an approach using Graph Convolutional Networks \cite{Kipf:2016tc} to model contextual cues, such as actor-actor and actor-object interactions, to improve action detection in video. On the challenging DALY dataset \cite{DBLP:journals/corr/WeinzaepfelMS16}, our model outperforms a baseline, which uses no context, by more than 2 points in Video-mAP, performing on par or better in all action classes but one. The learned adjacency matrix, visualized as an attention map, aids explainability by highlighting the learned context, such as objects relevant for recognizing the action, even in a zero-shot setting, i.e. for actions unseen during training. We quantitatively evaluate the attention maps using our proposed metric based on recall of objects retrieved by the attention. Results show the effectiveness of our model to highlight the relevant objects with high recall. All the above are achieved in a weakly-supervised setting using only up to five or even one actor box annotation per action instance. Future work includes end-to-end model training using weak supervision and modeling relations between consecutive clips and videos of the same action. 

\subsubsection*{Acknowledgements} We would like to thank Philippe Weinzaepfel for providing us with the predicted action tubes of their tracking-by-detection model.

\bibliographystyle{splncs04}
\bibliography{mybibliography}

\end{document}